\documentclass[letterpaper]{article} % DO NOT CHANGE THIS
\usepackage[]{aaai24}  % DO NOT CHANGE THIS
\usepackage{times}  % DO NOT CHANGE THIS
\usepackage{helvet}  % DO NOT CHANGE THIS
\usepackage{courier}  % DO NOT CHANGE THIS
\usepackage[hyphens]{url}  % DO NOT CHANGE THIS
\usepackage{graphicx} % DO NOT CHANGE THIS
\usepackage{natbib}  % DO NOT CHANGE THIS AND DO NOT ADD ANY OPTIONS TO IT
\usepackage{caption} % DO NOT CHANGE THIS AND DO NOT ADD ANY OPTIONS TO IT
\usepackage{algorithm}
\usepackage{algorithmic}
\usepackage{amsmath}

\usepackage{xcolor}
\newcommand{\answerYes}[1]{\textcolor{blue}{#1}} 
\newcommand{\answerNo}[1]{\textcolor{teal}{#1}} 
\newcommand{\answerNA}[1]{\textcolor{gray}{#1}}

\newcommand{\hl}[1]{\textcolor{black}{#1}} 

\usepackage{newfloat}
\usepackage{listings}
\usepackage{enumitem}
\DeclareCaptionStyle{ruled}{labelfont=normalfont,labelsep=colon,strut=off} % DO NOT CHANGE THIS
\floatstyle{ruled}
\newfloat{listing}{tb}{lst}{}
\floatname{listing}{Listing}

\title{Measuring Dimensions of Self-Presentation in Twitter Bios and their Links to Misinformation Sharing}
\author {
    Navid Madani\textsuperscript{\rm 1},
    Rabiraj Bandyopadhyay\textsuperscript{\rm 2},
    Michael Miller Yoder\textsuperscript{\rm 3},
    Stefan D. McCabe\textsuperscript{\rm 4},
    Briony Swire Thompson\textsuperscript{\rm 5},
    Kenneth Joseph\textsuperscript{\rm 1}
}
\affiliations {
    \textsuperscript{\rm 1} University at Buffalo\\
    \textsuperscript{\rm 2}GESIS Leibniz Institute for the Social Sciences\\
    \textsuperscript{\rm 3}University of Pittsburgh\\
    \textsuperscript{\rm 4}George Washington University \\
    \textsuperscript{\rm 5}Northeastern University\\
    smadani@buffalo.edu, rabiraj.bandyopadhyay@gesis.org, mmyoder@pitt.edu, stefanmccabe@gmail.com, b.swire-thompson@northeastern.edu, kjoseph@buffalo.edu
}

\begin{document}

\maketitle

\begin{abstract}

Social media platforms provide users with a profile description field, commonly known as a ``bio," where they can present themselves to the world. A growing literature shows that text in these bios can improve our understanding of online self-presentation and behavior, but existing work relies exclusively on keyword-based approaches to do so. We here propose and evaluate a suite of \hl{simple, effective, and theoretically motivated} approaches to embed bios in spaces that capture salient dimensions of social meaning, such as age and partisanship. We \hl{evaluate our methods on four tasks, showing that the strongest one out-performs several practical baselines.} We then show the utility of our method in helping understand associations between self-presentation and the sharing of URLs from low-quality news sites on Twitter\hl{, with a particular focus on explore the interactions between age and partisanship, and exploring the effects of self-presentations of religiosity}. Our work provides new tools to help computational social scientists make use of information in bios, and provides new insights into how misinformation sharing may be perceived on Twitter.
\end{abstract}

\section{Introduction}

\noindent On social media sites like Tumblr \cite{yoder2020phans}, Twitter \cite{liEmojiSelfIdentityTwitter2020,rogersUsingTwitterBios2021,pathak2021method}, TikTok \cite{darvinDesignResistancePerformance2022}, and Parler \cite{barFindingQsProfiling2023}, the profile description field (or \emph{bio}) asks users to describe themselves in a single text box. Users commonly do so with a combination of delimited phrases, each of which can range from unigrams like ``Republican'' to more complex expressions like ``2020 Election Truth Seeker'' \cite{pathak2021method}.  

\hl{In most social settings, people aim to present to others only a single social identity \cite{heise_self_2010,schroder2016modeling}}.  Examples of identities include social roles, like ``doctor’’ and ``mother’’, group memberships like ``Democrat’’ and ``Yankees fan’’, and social categories like ``black people’’ and ``women’’ \cite{tajfelIntegrativeTheoryIntergroup1979}. Bios \hl{thus} present a unique setting where we can observe individuals explicitly labeling themselves with multiple social identities \cite{marwickTweetHonestlyTweet2011a}.\footnote{Bios really express phrases that \emph{signal identity}, what \citet{pathak2021method} call \emph{personal identifiers}. Here, we retain the phrase (social) identity as a familiar and concise shorthand.} With respect to the language of self-presentation \cite{johnstoneLinguisticIndividualSelfexpression1996}, bios \hl{are therefore an important tool for social scientists for three reasons.} 

\hl{First, bios} provide insight into patterns of shared identity. For example, bios have been used to show that people who label themselves coffee snobs tend to be the same that label themselves marketing gurus \cite{pathak2021method}. \hl{They have also been used to better understand the increasing entanglement of partisanship and culture in social identities in the United States \cite{Essig2024,phillipsWhyPeopleThink2024}.}
\hl{Second, bios allow us to study links between the identities people select for themselves and behavior.} Users' choices on how to label themselves can be linked to, for example, decisions on who chooses to re-blog whom on Tumblr \cite{yoder2020phans}\hl{, or how \hl{people respond to marketing campaigns \cite{ngRecruitmentPromotionTwitter2023}.} Finally, bios can help us identify people who share a certain characteristic. To this end, p}rior work has also explored how bios can help identify individuals such as Qanon supporters \cite{barFindingQsProfiling2023} and journalists \cite{zeng2019detecting}, characterize the demographics associated with the self-presentation of particular identities \cite{pathak2021method}, and even reflect current social movements \cite{rogersUsingTwitterBios2021,hareSlavaUkrainiExploring2023}.

Existing quantitative work on bios has, however, has focused \hl{almost} exclusively on characterizing bios via the presence of particular phrases, whether through dictionary-based methods \hl{\cite{rogersUsingTwitterBios2021,zeng2019detecting,bar2023finding}} or through analysis of patterns in phrases extracted with regular expressions \hl{\cite{yoder2020phans,pathak2021method,Essig2024,ngRecruitmentPromotionTwitter2023}}.   While findings from these approaches can be illuminating, they also make it difficult to study how bios align with particular dimensions of importance to social scientists, such as partisanship \cite{kozlowskiGeometryCultureAnalyzing2019}.  \hl{Acknowledging these challenges and opportunties,  \citet{jiang2023retweet} develop an approach that uses modern NLP methods to infer ideological leanings implied by bios. However, their work focuses only on learning political ideology and also requires data beyond the bio itself to perform estimation. There thus exists a need for a method that can help us to measure self-presentation in bios in ways that 1) let us move beyond keyword-based methods and 2) beyond political ideology, 3) without requiring additional data.}

\hl{To this end, the goals of this paper are two-fold.} The first goal of the present work is to develop and evaluate three methods to project English-language social media bios onto \hl{multiple} dimensions \hl{of social meaning, without retraining a new embedding model for each new desired dimension}. \hl{Our methods are technically straightforward, in that we extend existing work on how to scale text along dimensions of meaning like gender and race using projections of text embeddings \cite{wilkerson2017large}. However, we build on this work by developing approaches specifically for the study of social media bios. Our models are \emph{grounded in the relevant social science}, in that each method we present draws on social identity theory \cite{tajfelIntegrativeTheoryIntergroup1979} to learn bio embeddings by} making use of the fact that many bios consist of multiple, clearly delimited social identities \cite{marwickTweetHonestlyTweet2011a,pathak2021method}. \hl{As such, unlike in standard embedding models where we aim to embed words that mean the same thing closely together, our models aim to create embedding spaces (and projections of them) where phrases that are \emph{applied to the same people} are close together in embedding space.}

After describing our three approaches, we present an extensive validation across \hl{four} tasks: 1) predicting which identities will appear in the same bio, 2) how projections of embeddings for specific identities within bios onto dimensions of age, gender, and partisanship correlate with human judgements, 3) whether projections using the embeddings of \emph{entire bios} also correlate with human judgements, \hl{and 4) whether projections using the embeddings of entire bios correlate with other measures of user ideology, and show similar associations to behavior}. Across all evaluation tasks, we find that the most effective model is an SBERT \cite{Reimers2019SentenceBERTSE} model \hl{fine-tuned on bios}. \hl{In Evaluation 1), we show that this model is best able to make predictions about identities the model has not seen, emphasizing its ability to generalize to new and unseen identities. Evaluations 2) and 3) show that this model effectively captures \emph{perceptions} of how individuals present themselves, which is critical for understanding how other users interpret a particular bio \cite{heiseAffectControlTheory1987,pathak2021method}. Finally, \emph{while our focus is on what is implied to others when a user self-presents, rather than inferring some ``truth'' about a user}, we show that estimates of user ideology by our method correlate with other known approaches for doing so. Moreover, we show that associations with behavioral patterns (of misinformation sharing) using our method are consistent with other approaches.}

Having identified a single model to embed bios, we \hl{continue to the second goal of our work, which is to explore} associations between self-presentation in bios and the rate of low- (relative to high-) quality news shares \hl{on Twitter}. \hl{Specifically, we focus on two research questions that have not been addressed in the existing literature.} \hl{First, perhaps the most well-established finding in this literature is that misinformation sharing is more prevalent among older, right-leaning individuals \cite{grinberg2019fake,guessLessYouThink2019,brashierAgingEraFake2020,osmundsenPartisanPolarizationPrimary2021,nikolovRightLeftPartisanship2021}. However, due in part to sample size restrictions, existing work has theorized \cite{grinberg2019fake} but not tested empirically the existence of an \emph{interaction} between age and partisanship. In the present work, using two different large datasets of Twitter users, we indeed show for the first time that such an interaction effect exists between self-presentation of age and  partisanship, in that self-presenting as older \emph{and} Republican has a multiplicative association with misinformation sharing. Second, while scholars have \emph{suggested} that religiosity is a critical dimensions of the self associated with misinformation sharing online, no empirical work has addressed this point \cite{druckmanRoleRaceReligion2021}. To this end, we present 1) new and convincing evidence that on average, presenting as more religious is strongly associated with misinformation sharing.}

In sum, the present work provides three contributions:
\begin{itemize}[noitemsep,nolistsep]
	\item We propose, evaluate, and make public\footnote{The model is publicly available for use on \url{https://github.com/navidmdn/identity_embedding}}
. a \hl{simple, effective, and theoretically motivated tool} to embed English-language bios in socially meaningful latent spaces. 
    \item We show that our method \hl{1)} can be used to project both individual social identities and entire social media bios onto salient dimensions of social meaning, such as partisanship, gender, and age in ways that correlate with human judgements in two new survey datasets, \hl{and 2) correlates well with other measures of user ideology}.
    \item We use our model to \hl{extend our understanding of the } relationship between how active news-sharing accounts on Twitter self-present and the proportion of news they share coming from low-quality news sites, \hl{particularly with respect to interactions between age and partisanship, and with respect to religiosity}.
\end{itemize}

%propose and evaluate three novel methods designed to measure the social dimensions of meaning expressed in Twitter bios. Our aim in doing so is to bridge the gap between individual social identities and the emergent social meanings in social media self-presentation. Specifically, we focus on Twitter as a platform due to its prominence as a social media platform for expressing diverse perspectives and engaging in public discourse.

\section{Background}

\paragraph{Measuring text on social dimensions of meaning} Social psychologists have developed a host of survey-based methods to measure associations between social identities and dimensions of social meaning. \hl{This estimation of identities on such dimensions is a core focus of social psychologists \cite{schroder2016modeling,fiske_model_2002}, because connecting self-presentation along specific dimensions of social relevance to social behavior is important for developing and testing new theories \cite{heise_self_2010}.} 
Below, we leverage these established approaches to evaluate our methodology. However, survey data do not scale to the myriad ways people identify themselves \cite{heise_self_2010}, are usually too small to capture differences across subgroups or contexts \cite{smith-lovinAffectControlAnalysis1992}, and struggle to account for linguistically complex identities or situations where multiple identities are applied \cite{joseph2021friend}. 

Most of the computational tools developed to address these challenges function by projecting embeddings from \hl{distributional semantic models (\emph{DSM}), such as BERT, onto particular dimensions of meaning.} The present work is most aligned with efforts that use contextualized embeddings \cite[e.g.][]{kuritaMeasuringBiasContextualized2019,lucyDiscoveringDifferencesRepresentation2022,fieldContextualAffectiveAnalysis2019a} to do so. \hl{However, we expand on these methods in that we aim to focus not on \emph{linguistic similarity}, but rather \emph{similarity in the types of people who use particular identities}. More specifically,} DSMs are based on the assumption that contextual similarity---similarity in where phrases appear in text---is a strong proxy for semantic similarity (roughly, synonomy). The idea behind this assumption is that phrases with high semantic similarity should have similar cognitive associations to other phrases, and thus high contextual similarity too \cite{millerContextualCorrelatesSemantic1991}. While deeply intertwined \cite{caliskanSocialBiasesWord2020}, these \emph{linguistic} associations and the \emph{socio-cultural} associations of interest to us differ. Linguistic associations represent phrases with similar associations to \emph{similar other phrases}; what we desire are phrases with similar associations to \emph{similar kinds of people}. \hl{While methodologically consistent with prior work, then, our work extends the existing literature by defining a different \emph{socio-theoretic} goal that is consistent with the existing literature on self-presentation.} We also provide several new evaluation datasets for future work.

In focusing on sets of identities applied to individuals, our work also relates to \emph{entity-centric} text analysis \cite{fieldEntityCentricContextualAffective2019}. Entity-centric work focuses on using phrases with known meanings (e.g. from surveys) to understand the portrayal of individuals  \cite{antoniakNarrativePathsNegotiation2019,mendelsohn2020framework}. Our method builds on a complementary idea, namely that we can use the fact that all identities in a bio refer to a single entity (a user) to create better embeddings. Similar in this vein is the work of \citet{bammanUnsupervisedDiscoveryBiographical2014c}, who use this idea to infer character personas in literature. The present work compliments these efforts by using entity-centric data to produce embeddings, rather than phrase clusters.

\hl{Our work also ties to the literature that explores the language associated with how people express misinformation \cite{mu2020identifying,shuFakeNewsDetection2017,rashkinTruthVaryingShades2017}. More specifically, we complement these efforts to study how people express specific instances of misinformation by exploring how people who tend to spread misinformation present themselves via particular (sets of) social identities. }

\paragraph{\hl{The Demographics (and Self-presentations) of Misinformation Sharing}}

The study of misinformation online has exploded in recent years \cite{lazer2018science}. Within this literature, several papers have looked at associations between demographics and rates of misinformation sharing. Typically, they do so by gathering demographics either via survey \cite[e.g.][]{grinberg2019fake} or using voter records \cite{guessLessYouThink2019,moslehSelfreportedWillingnessShare2020}, and then associating contained demographic information with rates of misinformation sharing.  Across studies, however, prior work has repeatedly found that the best predictors of misinformation exposure and sharing online are 1) old age \cite{brashierAgingEraFake2020}, 2) alignment with the political right \cite{osmundsenPartisanPolarizationPrimary2021,nikolovRightLeftPartisanship2021}, and 3) overall levels of online activity \cite{grinberg2019fake}. 

Our work compliments these existing efforts in \hl{a number of} ways. First, these prior works tend to use small-N samples of misinformation sharers because of the challenges and biases associated with their recruitment methods \cite{hughesUsingAdministrativeRecords2021}. For example, \citet{grinberg2019fake} and \citet{guessLessYouThink2019}, analyze sharing patterns in settings where only 400 Twitter users and 101 Facebook users shared any misinformation, respectively. In contrast, we analyze two different and much larger datasets; in the larger one, 77,190 accounts share at least one low-quality news link. 

\hl{Second,} we focus here on demographics \emph{conveyed through self-presentation}. These self-presented demographics are important in their own right for understanding who \emph{other} Twitter users \emph{perceive} to be sharing misinformation, perceptions that do not always align with a user's ``true demographics'' \cite{nguyen2014gender}.  
 \textbf{Critically, then, we do not claim that our method infers demographics of users, nor do we believe it does so.} Indeed, these self-presentations may vary from demographics in at least two ways. First, individuals may consciously choose not to convey certain demographic information online, such as gender \cite{bussTransgenderIdentityManagement2022}, and more generally choose which dimensions of the self are most important to present to their Twitter audience \cite{marwickTweetHonestlyTweet2011a}. Second, non-human accounts, such as those run by Russia's Internet Research Agency (IRA), may fabricate self-presentations to shape online discussion \cite{zhangAssemblingNetworksAudiences2021a}. Prior studies that link users to surveys or voter records rule out these latter accounts in their sampling approach. In contrast, we aim to focus on which dimensions of social meaning are salient, or cast as salient, by those sharing misinformation unknowingly or for manipulation.  We thus study here \emph{how the misinformation shares present their identity}.

\hl{These two distinctions shape the two novel research questions we study here. First, the larger size of our dataset lets us explore \emph{interactions between} self-presentations of age and partisanship, as compared to prior work which analyzes only the main effects. Second, our focus on dimensions of self-presentation instead of traditional demographics lets us analyze a novel, although well-theorized \cite{druckmanRoleRaceReligion2021}, empirical question: how is (self-presented) religiosity associated with misinformation sharing?}

\section{Methods for Embedding Twitter Bios}

We propose three models that leverage existing methods to project phrases onto dimensions of social meaning in different ways. Our first model uses data only from bios, whereas the latter two use fine-tuning to balance between meanings in bios and  semantic information in large, pre-trained DSMs. In all cases, our models are trained by using patterns in the multiple identities that appear in many social media bios. Because of this, it is useful to introduce some limited notation. First, let $X$ denote a dataset of bios where identities have been extracted, e.g. by using a regular expression 
 \cite{yoder2020phans,pathak2021method}. We assume 
 $X^i = \{x^i_1, x^i_2, ..., x^i_k\}$ represents a set of $k$ identities extracted from a single bio, and that $V$ is a vocabulary of all unique identities in the training portion of $X$.

\subsection{Models}

\paragraph{Bio-only model}
Our \emph{Bio-only model} is constructed by applying \texttt{word2vec} \cite{mikolov_efficient_2013} to $X$.
In common terminology for \texttt{word2vec}, we treat identities as words and bios as a context. Our intuition is that if the \texttt{word2vec} model can leverage contextual similarity on the ``word-to-linguistic context'' matrix to identify words with shared semantic meanings, it may also be useful to leverage the ``identity-to-person context'' matrix to identify phrases with shared socio-cultural meanings. We use \texttt{word2vec} models with an embedding size of 768 to match the embedding size of the other models used below, and train for 300 epochs with a window size of 8 (only .01\% of bios in our training data contain more than 8 identities). Additional minor details are provided in the appendix.

\paragraph{Fine-tuned BERT} To fine-tune BERT, we use a masked language modeling (MLM) objective, randomly masking one of the identities in each bio. %This approach was based on initial findings that forcing the model to predict full identities generated better embeddings in terms of ad hoc nearest neighbor queries than the standard approach to MLM. 
 To prepare our dataset for training, we take each of the instances $X^i$ and concatenate the phrases in it to form a full sentence. We then mask one of the identities and fine-tune a \texttt{BERT-base} model for 5 epochs while monitoring 10\% of the training set as validation data.   We used a learning rate of 2e-5 with a batch size of 64. Model training took approximately one day using a single A100 GPU. \hl{When bios are shorter than the context window size, padding is added; attention masking is used to ensure padding does not impact the embedding. We keep the embedding size of the BERT model at the default 768.}

\paragraph{Fine-tuned SBERT} Finally, we construct a final model by fine-tuning Sentence-BERT  \cite{Reimers2019SentenceBERTSE}. 
Sentence-BERT uses \emph{contrastive learning}, where the learning setup must be carefully constructed \cite{Schroff2015FaceNetAU}. We develop an intuitive but effective approach here based on shared identities. In a contrastive learning framework, each data point is a \emph{triplet} consisting of an \emph{anchor}, a \emph{positive}, and a \emph{negative} sample. Our goal is to reshape the embedding space through fine-tuning such that for each triplet, the distance between anchor and positive samples, which co-occur within a bio in our setup, is minimized while the distance between anchor and negative samples, which do not, is maximized. \hl{As above, padding and attention masks are used when bios are shorter than the size of the context window. The embedding size of our selected sentence BERT model is 768 which matches the embedding size of all other proposed approaches for fair comparison.}

We can frame this contrastive learning problem as a regression task: given a triplet of anchor ($X_a$), positive ($X_p$) and negative ($X_n$) samples and a similarity measure (here, cosine), our objective is for $cs(X_a, X_p) = 1.0$ and $cs(X_a, X_n) = 0.0$ for all training points, where $cs$ stands for cosine similarity. We can then optimize this objective using mean squared error. The challenge is to construct an effective set of triplets to train on. To do so, we first take a bio $X^i$ from the training set, and then randomly select an identity from $X^i$ to be the positive sample. We name the remaining identities in $X^i$ the anchor sample. Finally, we randomly select an identity that never co-occurs with the positive sample as the negative sample. As an example, from the bio \emph{[assistant professor, Bernie supporter, \#blacklivesmatter]} we set \emph{assistant professor, \#blacklivesmatter} as the anchor sample, \emph{Bernie supporter} as the positive sample, and randomly select a negative sample that never co-occurred with \emph{Bernie supporter}. We construct a triplet for each $X^i \in X$ using this method, and use these to fine-tune an \emph{mpnet-base} Sentence-Bert model.  Models were trained for 5 epochs for one day on a single A100 GPU. 

\subsection{Training Data}\label{sec:training_data}

All models are trained on a sample of 3,534,903 bios from users who sent an English-language tweet captured in the Twitter Decahose\footnote{A sample of approximately 10\% of all tweets.} in 2020.  We use the method from \citet{pathak2021method} to extract identities from bios. Their extraction method consists of two steps, one where bios are split into chunks using a manually crafted regular expression, and a second cleaning step. For example, from the Twitter bio ``Progressive Christian, wife, I am a proud Canadian," their method extracts \emph{Progressive Christian}, \emph{wife} and \emph{proud Canadian}.  Given that our models learn from patterns in shared identity, we use only bios that contain at least two identities for training, with $|V|$=22,516. A complete description of the dataset is available in the appendix.

\section{Evaluation}\label{sec:eval}\label{sec:eval_predict}

We conduct \hl{four} evaluations to assess model validity. The first is a prediction task, where we evaluate the full embedding space of each model and its ability to capture information that shapes users' decisions on which sets of identities to place in their bio.  The second connects our work to the existing literature on embedding and projecting individual social identities onto dimensions of social meaning. The third assesses our ability to project entire bios onto meaningful dimensions, with an eye towards validating our approach for our case study. \hl{Finally, the fourth task shows that our measures of partisanship correlate with other existing measures, and that our methods are associated with the behavior of interest (misinformation sharing) in the same way.} Here, we describe each task and results for it in \hl{four} separate subsections. 

%  While the projection of individual phrases is of use to social psychologists who are interested in studying the links between identity, social meaning, and behavior \cite{heiseAffectControlTheory1987,cuddyBIASMapBehaviors2007}, linking self-presentation to behavior online requires that we understand what a complete bio conveys.

\subsection{Can we Predict Held-out Identities?}

\paragraph{Task Description}  
To perform this evaluation, we randomly sample another 1.5M Twitter bios from the Decahose using the same approach as above, creating an 80/20 train/test split between the two samples. For each observation in the test set, we ensure that \emph{at least one} of the identities is in $V$ (which is defined using the training data). For the $i$th sample, we then take one identity, $X^i_t$, as the hold-out target and call the rest of the bio $X^i_r$. We ensure  $X^i_t$ is in $V$, i.e. in all cases the target is observed at least once in the training data.  This task notably favors the three new models we present here, in that they are each trained in a manner similar to the prediction task. However, the comparison is still useful to 1) ensure that fine-tuning works as expected and 2) to compare between the three new models. With respect to the latter point, there is reason to believe that the word2vec model should outperform the Fine-tuned BERT or fine-tuned SBERT model, because the word2vec model is trained \emph{only} on in-domain data relevant to the task.

To generate predictions, we first generate an embedding for $X^i_r$, $ L^i_r = embedding(X^i_r)$\footnote{Extended details on how embeddings are generated for each model are in the Appendix.}. We then measure the \hl{cosine} similarity of $L^i_r$ with the embedding of all identities $v \in V$, $Similarity(L^i_r, L^i_v)$, leaving us with $|V|$ similarity scores to $L^i_r$.%\footnote{Recent work has suggested that transforming embeddings before similarity comparisons is important \cite{timkey-van-schijndel-2021-bark}. We do not find performance improvements here, and so do not do so.}  
We evaluate similarity scores returned by each model using two evaluation metrics: \emph{average rank} and \emph{log softmax score}. Average rank is computed by, for each test point, finding the ranking of $X^i_t$ in the scores produced by each model, and taking the average over all test points. The log softmax score draws on prior work \cite{joseph2021friend} and transforms similarity scores into a probability distribution using the softmax, and then takes the log of the result for $X^i_t$.

Finally, for evaluating the out-of-domain generalizability of our models, \hl{i.e. how well the model generalizes to unseen identities,} we split our test data into two sets, a \emph{Main Evaluation} set, where $X^i_r$ also contains at least one identity observed in the training data, and a \emph{Generalizability} set, in which no identities in $X^i_r$ are seen in the training data. This is necessary to fairly evaluate our \emph{Bio-only model}, which has a restricted vocabulary and does not generalize to out-of-domain phrases, to the other models, each of which are capable of handling out-of-domain text.\footnote{Note that $X^i_r$  can still contain phrases that the Bio-only model does not recognize and replaces them with the 0-vector.} It is also a useful test of the (in-domain) generalizability of the other models. We evaluate results separately for these two test datasets.

We compare our models to three baseline DSMs used frequently in prior work: BERT \cite{Devlin2019BERTPO}, RoBERTa \cite{liu2019roberta}, and Sentence-BERT \cite{Reimers2019SentenceBERTSE} \hl{(see appendix for details)}. As in much of the prior work \cite{lucyDiscoveringDifferencesRepresentation2022}, these approaches are \emph{not} fine-tuned on bios, giving a baseline for how important in-domain training is for our problem. 

\paragraph{Results} 
        \begin{figure}[t]
            \centering
            \includegraphics[width=1\linewidth]{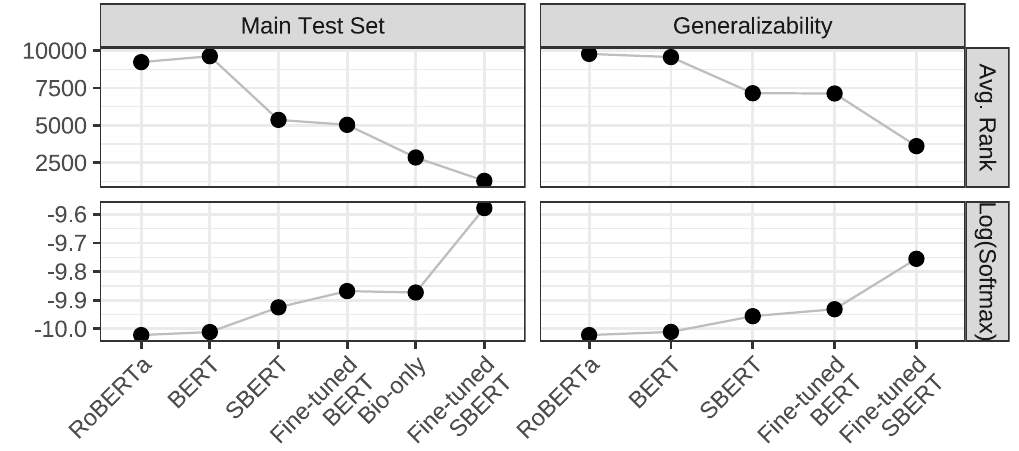}
            \caption{
                Performance of each model (x-axis) on each of our two outcome metrics (separate plot rows) for the Main Test set and the generalizability test set (separate plot columns). Note that for rankings, lower is better.
            }
            \label{fig:performance}
        \end{figure}
    
In contrast to our expectations, the Fine-tuned SBERT model consistently outperforms all other models on both evaluation metrics on the Main Test set.  Figure~\ref{fig:performance} shows this, and also reveals that the next best model, in all cases, was the \emph{Bio-only model}, and that the \emph{Fine-tuned BERT} model does not show the same jump in performance relative to the baseline BERT model that the \emph{Fine-tuned SBERT} model does. Finally, we see that the baseline SBERT model outperforms the baseline BERT model. 

The fact that the fine-tuned SBERT model improves over the Bio-only model, but the Fine-tuned BERT model does not, is evidence that knowledge from the pre-trained SBERT model (but not BERT) is useful in our setting.  It also shows that performance gains cannot only be attributed to fine-tuning on in-domain language, but instead that our contrastive learning setup was effective and that Sentence-BERT is indeed the more effective initial model for fine-tuning, at least on this evaluation task. The second column of Figure~\ref{fig:performance} shows that the Fine-tuned SBERT model also performs best on the Generalizability test set. Even when the Fine-tuned model is not exposed to any of the identities in $X^i_r$, it improves by nearly 100\% over the standard SBERT model in terms of average rank.  Figure~\ref{fig:performance} also makes clear that there is room for improvement. To this end, we conduct an error analysis\hl{; see the appendix for details.}

 %At a high level, our error analysis compares predictions from the Fine-tuned SBERT and the Bio-only model and shows most interestingly that the gains of the SBERT model over the Bio-only model indeed come from the ability to capture semantic meaning in compositionl identities (e.g. not  %had a bimodal ranking distribution, with varying levels of performance for different test points. It  struggled to learn from compositional identities and failed to leverage relevant external knowledge effectively. In contrast, the Fine-tuned SBERT model demonstrated superior performance by correctly ranking identities, leveraging contextual cues such as gender stereotypes and utilizing external knowledge.

\subsection{Do Projections of Single Identities Correlate with Human Judgements?} \label{sec:eval_single}

\begin{figure}
    \centering
    \includegraphics[width=\linewidth]{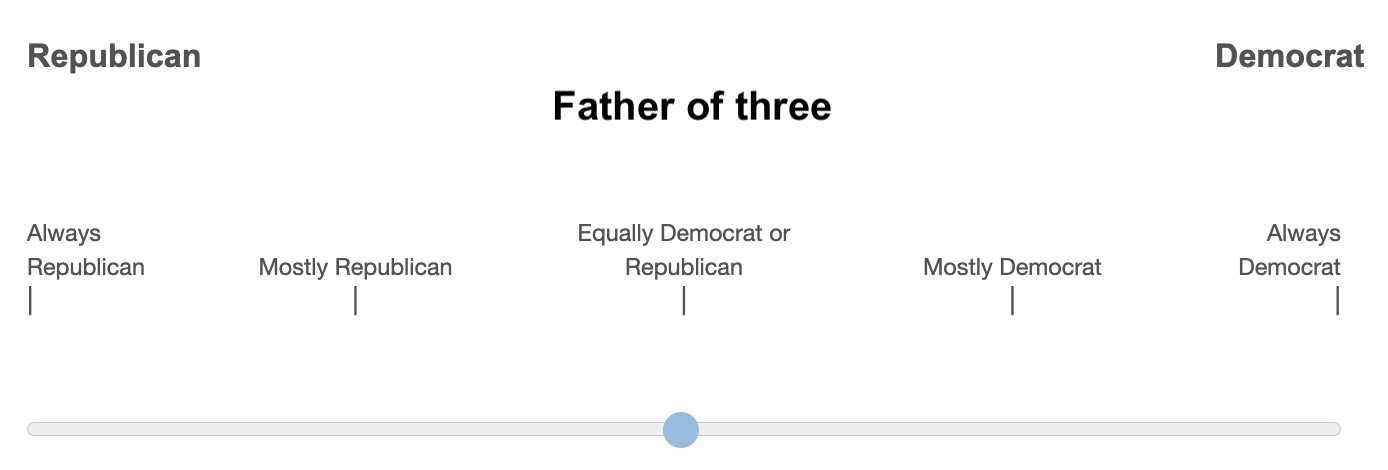}
    \caption{An example of a survey question asked on the single identity projection evaluation for the identity ``father of three'' on the partisanship dimension}
    \label{fig:slider_identity}
\end{figure}

\paragraph{Task Description} Following prior work \cite{joseph2020word}, we assess how well our embeddings can be used to project identities onto dimensions of social meaning in ways that correlate with human judgements. As no prior work focuses on identities that are common in bios, however, we construct a novel survey dataset for this task.  We provide brief details on the study here, full details can be found in the Appendix.

We asked 140 respondents on Prolific to rate 250 common identities in Twitter bios on three widely studied dimensions: gender, age, and partisanship.  We use the mean of ratings for each identity on each dimension in our analysis.  To select identities, we ranked identities in the Twitter dataset by frequency, and then manually selected the first 250 phrases that clearly signaled identity. For each identity on each dimension, respondents were asked to move a slider to represent their perception of where people who label themselves as that identity were likely to fall. For partisanship, for example, the slider ranged from ``Always [a] Democrat'' to ``Always [a] Republican.'' Figure~\ref{fig:slider_identity} provides an example question. 
 For gender and age, we followed the approach outlined by \citet{joseph2020word} exactly, using the same slider.  Specifically, for age, participants were asked to rate identities on perceived age from 0-100, for gender, the question appears as in Figure~\ref{fig:slider_identity}, replacing ``Republican'' with ``Man'' and ``Democrat'' with ``Woman.''

To construct projections for each embedding model, for each identity, onto these dimensions, we follow the literature and 1) embed identities as defined above, 2) define a set of words and phrases that denotatively characterize each ``end'' of the dimension (e.g. ``man'' vs. ``woman'' for gender) and then 3) use these to project each identity onto a line in the embedding space defined by those two dimension ends, giving a single number. Several approaches exist to complete steps 2) and 3) \hl{\cite{joseph2020word}}. We follow prior work where possible for 2), and for 3), \hl{i.e. to calculate similarity, we use RIPA, the method described by  \citet{ethayarajhUnderstandingUndesirableWord2019}}. Finally, for each embedding model on each dimension, we then compute the Spearman correlation between the projections and the survey data.

\paragraph{Results}

\begin{figure}
    \centering
    \includegraphics[width=1\linewidth]{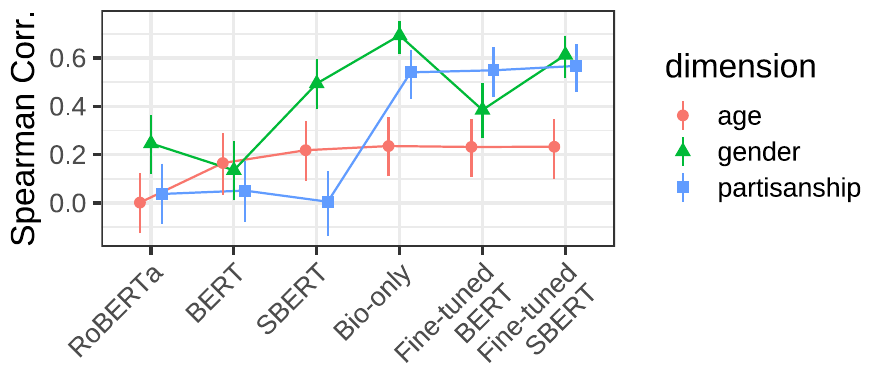}
    \caption{The Spearman correlation (y-axis) between projections and human judgements on 250 social identities for each model (x-axis) and dimension (shape/color). Error bars are 95\% bootstrapped confidence intervals.}
    \label{fig:eval2}
\end{figure}

Figure~\ref{fig:eval2}  shows that our three models all perform at least as well---but not better than---the baselines for projections onto age, all significantly improve over baselines in measuring perceived partisanship, and significantly improve over the baselines, except for Fine-tuned BERT, on perceived gender. Our models' improvements over baselines are particularly salient in comparison to the RoBERTa and BERT baselines. Notably, we also see that our Bio-only model, based on word2vec performs well compared to the more complex models on this task. This suggests that for phrases that are prominent in bios, this model (which trains only on the bios themselves) may be preferred. However, the Bio-only model cannot extend beyond the vocabulary, and thus (as we will see) struggles with generalizability. Finally, Figure~\ref{fig:eval2} also shows that the greatest jumps in performance for our models, relative to the baseline, are clearly on the partisan dimension. 
 %Given the established salience of politics on social media relative to other realms of social life \cite{bailExposureOpposingViews2018}, our work appears to further validate claims \cite{joseph2020word} that connections exist between how well projections of embeddings capture particular dimensions of social meaning and how salient those dimensions are  in the training data.% This point is further enforced by the fact that performance improvements on domain-relevant identities do \emph{not} extend to out-of-domain identities. On the more traditional identities in the survey data from \citet{bolukbasiManComputerProgrammer2016b}, the non-fine-tuned SBERT model actually performs significantly \emph{better} (Spearman correlation 0.60, [0.54,0.65] than the best in-domain model, Fine-tuned SBERT (0.52, [0.45,.57]).

% As suggested elsewhere \cite{lucyDiscoveringDifferencesRepresentation2022,joseph2020word,fieldEntityCentricContextualAffective2019}, then, it would appear that the utility of semantic knowledge from pre-trained models is limited for identifying stereotypes of salient identities on salient dimensions of meaning for domain relevant identities.

\subsection{Do Projections of Entire Bios Correlate with Human Judgements? } \label{sec:eval_bio}

\paragraph{Task Description} \hl{Our third} evaluation \emph{assess the question,} can our model capture perceptions of self-presentation in entire bios? To address this, we conduct a similar analysis as above, comparing the Spearman correlation of projections of our embedding models to mean ratings by Prolific respondents on a survey task. There are, however, four primary differences between the evaluation of single identities and the one presented here for full bios.

First, of course, is that we ask respondents for their perceptions of entire bios, rather than individual identities within bios.  More specifically, we randomly sample 1,300 bios of users in our case study data, described below. Second, because in our case study we are interested in religiosity as well as age, gender, and partisanship, we add a question regarding perceived level of religiosity to the survey. Third, as opposed to selecting only one approach for defining ends of the semantic axis onto which embeddings are projected, we consider two approaches. The first follows our second evaluation and uses a combination of prior work and author intuition to define the terms at each end of the axis. Motivated to ensure accurate measurements for our case study, the second considers whether or not we can improve correlations with human judgement by constructing lists of terms for dimension endpoints that are informed by a qualitative analysis of bios. More specifically, we manually explore bios from the case study data described below that are not included in the survey study and use them to define the list of terms.  We compare performance using both approaches here, \emph{but ensure that we compute results only on bios that do not include terms that explicitly define our endpoints}. Fourth, we opt to only compare performance of SBERT and Fine-tuned SBERT, as other models fared poorly on one or both of the prior evaluations.  

All other details of our evaluation generally match those in our second evaluation, save for the sample of Prolific users; \hl{see the appendix for details on this.}
\paragraph{Results}\label{sec:results}

    \begin{figure}[t]
            \centering
            \includegraphics[width=1\linewidth]{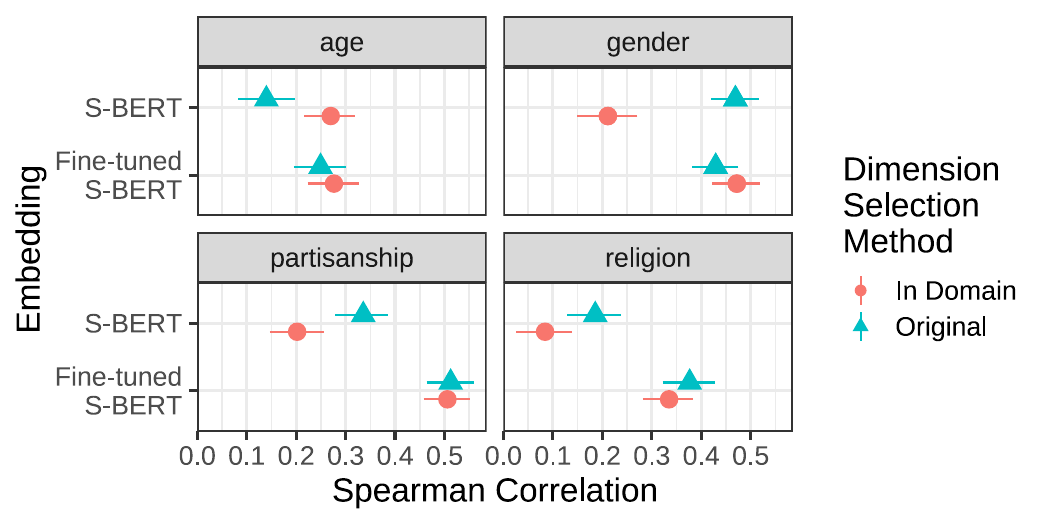}
            \caption{Spearman correlations (x-axis) between projections onto four different social dimensions (separate subplots) for SBERT and Fine-tuned SBERT (y-axis) for two different dimension selection methods (color/shape). Error bars are 95\% bootstrapped CIs.}
            \label{fig:eval3}
        \end{figure}

Figure~\ref{fig:eval3} shows that the Fine-tuned SBERT model has a significantly higher correlation with human judgements  on dimensions of partisanship and religion, regardless of how endpoints of the dimensions are defined. For gender and age, Fine-tuned SBERT performance shows no significant improvement over a non-fine-tuned SBERT model, but is also not significantly impacted by how dimension endpoints are defined. In contrast, for the SBERT model, in one case (age) our use of qualitative exploration to find in-domain sets of phrases to define dimensions significantly increases correlations with human judgement, and in the other (gender) it significantly decreases correlations. These results suggest that the Fine-tuned SBERT model is therefore more robust to the known challenge of defining ``good'' sets of phrases to define endpoints \cite{joseph2020word}.

\subsection{\hl{How do our Projections Compare to Other Measures at the User Level?}}

\paragraph{Data}

\hl{In our final evaluation, we focus specifically on projections of bios onto partisanship and explore 1) how well these projections correlate with a different measure of user partisanship (evaluating convergent validity), and 2) whether or not our measure of partisanship correlates in expected ways---and in a similar way to the other measure of user partisanship--- with misinformation sharing \cite[evaluating hypothesis validity][]{jacobs2021measurement}.  To conduct these evaluations, we make use of a pre-existing dataset of 374,684 Twitter users 1) for whom prior work \cite{mccabeNewTweetScoresDid2022} has computed a standard measure of user ideology using user follower networks, and 2) for whom we have data on the sharing of low-quality news sites.}

\hl{With respect to the partisanship measure we compare to, \citet{mccabeNewTweetScoresDid2022} apply a method pioneered by \citet{barberaTweetingLeftRight2015a} that uses a combination of information about the partisanship of elite Twitter users and follower relationships to compute an estimate of user partisan ideology. \citet{mccabeNewTweetScoresDid2022} adopt the same approach, but update information about elites from the 2012 data used by  \citet{barberaTweetingLeftRight2015a} to 2020 data, and apply the approach to a large set of Twitter users that they linked to voter registration records using established methods \cite{hughesUsingAdministrativeRecords2021}. The work from \citet{mccabeNewTweetScoresDid2022} presents full details on implementation and an extensive validation of the measure in comparison to voter registration data; we therefore refer the reader to their work these details and focus here only on a comparison between their updated version of \citeauthor{barberaTweetingLeftRight2015a}'s \citeyear{barberaTweetingLeftRight2015a} method and our partisanship measure using bio data.}

\hl{With respect to the sharing of low-quality news, we focus on the proportion of URLs to low- and high-quality news websites shared on Twitter by these users from July 1st, 2020 through May 31st, 2021. We determine whether a URL comes from a low- or high-quality news website by making use of 1) the list produced by \citet{grinberg2019fake} (who categorize sites into a binary  high or low-quality distinction) and 2) the NewsGuard domain rating list. NewsGuard is an organization that maintains a widely-used \cite[e.g][]{altayQuantifyingInfodemicPeople2022,horneDifferentSpiralsSameness2019} list of news websites that are rated on a scale of 0-100 for information quality. We follow prior work and dichotomize their ratings into a binary low- or high-quality value for each site. We use a score of 60 as the threshold, following prior work \cite{lin2022high}. Further, while we opt for a particular definition across these two lists, we note that prior work \cite{lin2022high} has shown that various lists of misinformation domains are highly correlated, and thus that results for the study of misinformation are unlikely to be sensitive to these types of changes.  Using these binary domain scores, we define our outcome variable for an individual user as the percentage of all URLs the user shares that come from a low-quality news site, divided by the total number of URLs shared from a domain listed in NewsGuard. We refer to this outcome as the \emph{proportion of low-quality shares}.}

\hl{Finally, for our bio-based measure, we use user bios collected in July of 2020 (i.e. within the same timeframe of the following data and misinformation sharing data), and project bios onto partisanship using the methods detailed in our evaluation of full bio embeddings (our third evaluation).}
 
\paragraph{Results}

\hl{We find a correlation of 0.39 [0.387,0.393] between the bio-based projection measure of partisanship using our fine-tuned SBERT model and the partisanship estimates from \citet{mccabeNewTweetScoresDid2022}. This correlation is nearly double the correlation we find between the bio-based projection measure computed using the non-fine-tuned SBERT model and the estimates from \citet{mccabeNewTweetScoresDid2022} (0.197 [0.193,0.201]), and even higher than the correlation between the fine-tuned SBERT model and the non-fine-tuned model (0.347 [0.343,0.35]). This strong correlation, even when compared to reasonable baselines, gives us further confidence in the convergent validity of our measure---while we should not expect a measure based on a more private behavior (following) to correlate exactly with one based on a more public behavior (text in bios), we do see, as expected, a reasonably strong relationship.}

\begin{figure}[t]
    \centering
    \includegraphics[width=\linewidth]{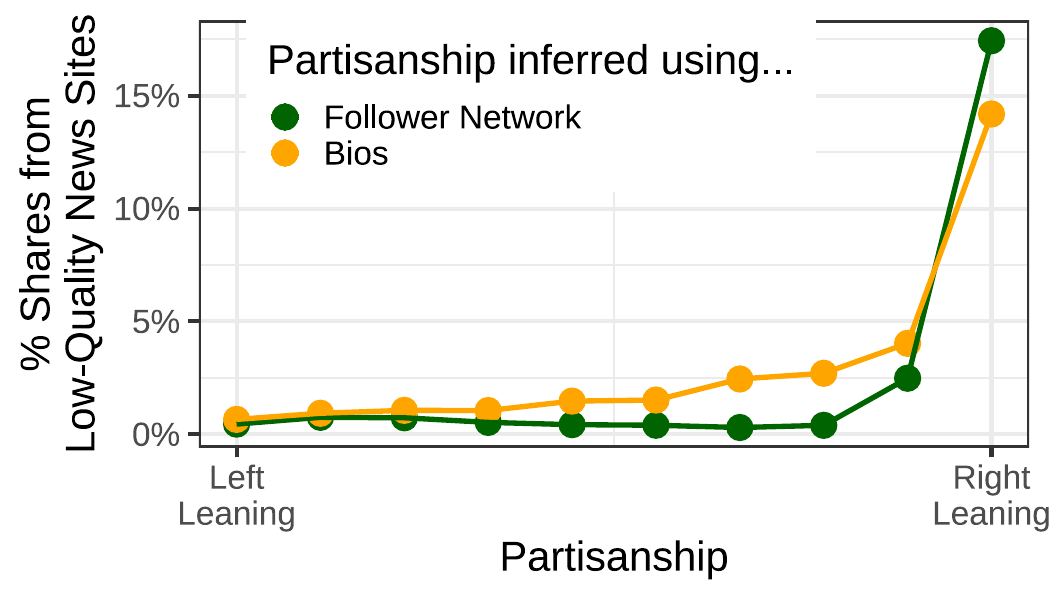}
    \caption{\hl{Proportion of low-quality shares (y-axis) across bins (n=10) of projections onto the partisanship (x-axis) dimension, estimated using two different methods (color). Error bars, while small, are present in the figure, and represent 95\% normal CIs.}}
    \label{fig:partisan-compare}
\end{figure}

\hl{We also find that these two measures show consistent estimates of the relationship between partisanship and misinformation sharing. These estimates, moreover, conform with what we would expect from prior work \cite{grinberg2019fake,guessLessYouThink2019}. More specifically, Figure~\ref{fig:partisan-compare} shows the estimated proportion of low-quality shares from a binomial regression model where the independent variables are indicators for partisanship after bucketing each variable into 10 separate bins of equal numbers of users, respectively.\footnote{Binning is done to avoid assumptions about the precision of these proxy variables and to ease interpretation; bin sizes selected here are for visual clarity. Results are not, however, sensitive to this decision.} All main effects in the model used to estimate the proportions shown in Figure~\ref{fig:partisan-compare} are significant at $p\leq.001$, and trends between the two different measures are consistent in showing that those furthest to the partisan right are most likely to share  low quality news. }

\section{Misinformation Case Study}

% To this point, we have shown that our novel identity-based embedding approach can better predict which sets of phrases will co-occur in bios, and better project individual identities onto important dimensions of social meaning. Two additional questions remain. First,  Second, assuming we can effectively capture these meanings of entire bios, \emph{what can we actually use these projections to study?}
% In this section, we address these two questions with a

The previous evaluations show that our Fine-tuned SBERT model can project bios onto social dimensions of meaning in ways that correlate with how those bios are likely to be perceived by humans\hl{, and that the approach has both convergent validity and hypothesis validity in comparison to a different measure of user partisanship. Our case study uses the same dataset, but leverages our methodology to explore the link between self-presentation in bios and the rate of low- versus high-quality news sharing in the context of our two research questions. First, we explore the \emph{interaction between} self-presentations of age and partisanship and their association with low-quality new sharing. Second, we consider the association between self-presenting as religious and misinformation sharing.}

\hl{We note that while it would be possible for \citet{mccabeNewTweetScoresDid2022} to make use of additional voter registration data to, e.g., control for age, our tool is still useful for two reasons. First, most existing social media datasets are not connected to voter registration data, nor are the aligned with follower network data that could be used to apply the method from \citet{barberaTweetingLeftRight2015a}. To this end, we are able to use our methods to replicate our analysis on a new dataset from the Twitter Decahose, which we could not do otherwise (see appendix for details). Second, we note that an analysis of religiosity is in any case not possible with existing voter registration data, as religiosity does not exist within versions of these data that have been provided to researchers in the past.}

\subsection{Results}

        \begin{figure}[t]
            \centering
            
            \begin{tabular}{c}
                \includegraphics[width=\linewidth]{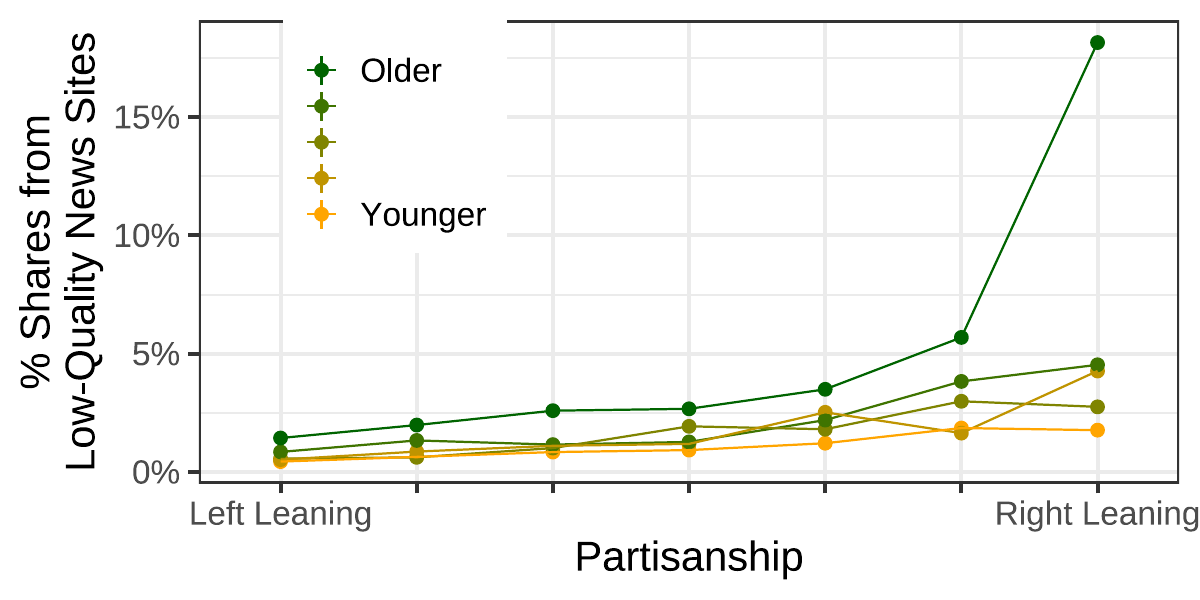} \\
                \includegraphics[width=\linewidth]{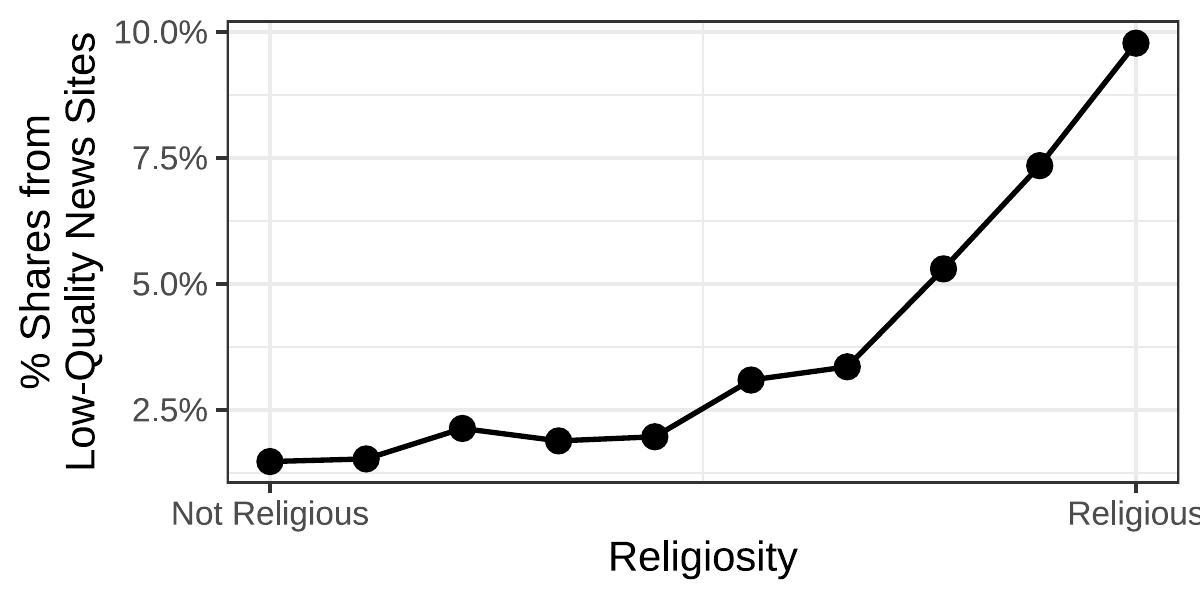} \\
            \end{tabular}
            \caption{A), Top: Proportion of low-quality shares (y-axis) across bins (n=10) of projections onto the partisanship (x-axis) and  projections onto the age (n=5 bins) (color) dimensions. Error bars, while small, are present in the figure, and represent 95\% normal CIs.\newline B), Bottom: Univariate model outcomes for religiosity}
            \label{fig:age_partisanship}
        \end{figure}

With respect to our first research question on the interaction between age and partisan self-presentation, we find that the direction of effects for age and partisanship generally match prior work, but that interaction effects not observed in prior work on demographics appear in our analyses. Figure~\ref{fig:age_partisanship} more specifically shows that users who self-present as older \emph{and} right leaning is associated with a superlinear increase in proportion of low-quality news shared relative to being old or right-leaning. 

Addressing our second research question, and using the same binning approach with univariate statistical models, we observe in Figure~\ref{fig:age_partisanship}B) that self-presenting as being highly religious is also associated with statistically significant ($p<.001$) jumps in the proportion of low-quality shares. As in the real world \cite{druckmanRoleRaceReligion2021}, our measures of religiosity and partisanship are highly correlated \emph{(at 0.76)}, and thus we do not attempt to tease out whether this effect is particularly salient conditioned on partisanship.  More specifically, we avoid interpretation of coefficients in a model with both of these variables because interpretation of regression coefficients with strongly co-linear variables is prone to misidentification of effect size and statistical significance \cite{gregorich2021regression}.

We do find, however, that religiosity adds additional predictive power, in that adding religiosity as a predictor to a model with partisanship results in a statistically significant increase ($p<.001$) in predictive power.  Descriptive statistics do, however, provide some evidence into a potential interactive relationship between partisanship, religiosity, and sharing content from low-quality sites. Specifically, Figure~\ref{fig:case_part_relig} in the appendix shows that partisanship and religiosity may have a multiplicative association with sharing content from low-quality news sites. Participants who present as both right-leaning and religious have, on average, higher odds of sharing low-quality news links than users who present as only right-leaning or only religious. Future experimental work targeting this specific interaction effect in ways that avoid endogeneity concerns might therefore be of interest.

\hl{Critically, all findings presented here extend beyond this particular dataset to a second large dataset of misinformation sharing drawn from a broader population in the Twitter Decahose. These replication results are presented in the appendix.}
 
\section{Discussion and Conclusion}

We present, evaluate, make public, and make use of new methods to project Twitter bios onto meaningful social dimensions. \hl{Methodologically, our contribution is not innovation in algorithm development, but in the application of standard algorithms to carefully constructed training datasets and training set-ups to build entity-centric identity embedding spaces (where identities that are commonly presented for the same people are represented in a similar space).} Notably, our estimates of self-presentation do not use other user behaviors (e.g. retweets). %While we might better estimate user information by looking at other behaviors as well, core questions in the social sciences are interested in the connection between self presentation and social identity (which we focus on measuring here) and other behaviors. Because our measure is focused \emph{only on the bio}, we can use its outputs to study how self presentation is associated with these other behaviors, rather than combining them together to predict other things. 

\hl{After extensively validating our proposed methods, we present a case study on two open questions in the literature on misinformation sharing online.  In response to our first question, we find an interaction effect between age and partisanship- that Twitter users presentinand Republican share a much higher proportion of low-quality news. With respect to our second research question, we similarly find that individuals who self-present as religious, perhaps especially on the political right, are much more likely to share a higher proportion of low-quality news. Our results have two important implications. With respect to combating misinformation online, while much work has considered interventions targeted along partisan \cite{martelEfficacyAccuracyPrompts2024} \emph{or} age-based \cite{brashierAgingEraFake2020}, our results provide evidence that we should be considering even more narrow interventions targeted at older \emph{and} right-leaning Americans. With respect to better understanding the misinformation environment online in the U.S., our work provides critical empirical validation of theories that suggest religiosity is an important component of the quality of information consumed and spread online \cite{druckmanRoleRaceReligion2021}.}

However, our work contains a number of limitations that should be considered. \hl{Specific to our case study, we follow previous work that uses a domain-based approach to identify misinformation sharing \cite{grinberg2019fake,guessLessYouThink2019,mooreExposureUntrustworthyWebsites2023a}, which includes posts that share a URL with a negative annotation. However, like previous work, we find the number of such posts to be a small proportion of all shares.} While other work referenced above suggests our efforts should extend to other social media sites, Twitter, like all sites, has particular elements that limit our ability to generalize claims about self-presentation elsewhere, in particular to ``offline'' behavior settings   
\cite{devitoTooGayFacebook2018}.
In particular, our models learn cultural associations from the majority white, male, younger, and left-leaning users of Twitter \cite{hughesUsingAdministrativeRecords2021}. Similarly, our survey evaluation data is largely drawn from white, American women, which limits their generalizability.  

 \hl{Our work is also contextualized in a particular period of time, and thus may not be responsive to significant shifts in social identity.} All of these, and results described in the paper, emphasize the contextualized nature of our trained models and datasets. While we hope that future work might expand beyond them, others should consider using our publicly available code to fine-tune our models on their own data. Finally, our case study is necessarily terse, and does not explore a number of additional important and interesting questions. \hl{In particular, there may be other interesting dimensions of self-presentation that may be interesting to social scientists interested in misinformation sharing on social media.} Our work also has ethical implications, which we discuss in more detail in the Ethics Statement below, as well as in the required paper checklist. Our hope, however, is that the proposed, and public, methods allow future efforts that explore new questions that link identity and behavior on Twitter and other sites with social media bios.

 \section{Acknowledgements}

 NM and KJ were supported by an ONR MURI N00014-20-S-F003 and by NSF IIS2145051. SM was supported by the John S. and James L. Knight Foundation through a grant to the Institute for Data, Democracy, \& Politics at the George Washington University. RB was supported by the MSCA Grant Agreement No. 101073351 at GESIS Leibniz Institute for the Social Sciences.

{
\fontsize{9.8pt}{10.8pt}
\selectfont
% \bibliography{main}

}

\section{Paper Checklist}

One can crudely separate the way that members of the NLP community acknowledge sociocultural associations embedded in language into three camps. In the first, acknowledgement is largely non-existent, with scholars either choosing to ignore the social implications of language or arguing that it is not within their purview to address them. In the second, scholars concerned with the potential that NLP tools might reify biases in language have sought to develop tools to reduce or remove the use of biased tools in production. Example efforts include the literature on ``debiasing'' embedding spaces---where scholars aim to remove meaning on ``unwanted dimensions'' (i.e. on social dimensions of meaning) and keep it only on ``objective'' dimensions (i.e. on more traditional semantic dimensions)--- and work to carefully elaborate how problems can arise from NLP tools that do not critically engage with the social implications of language \cite{blodgettLanguageTechnologyPower2020,field-etal-2021-survey}. 

While we strongly support the latter line of critical scholarship, we fall into a third camp of NLP researchers, who see the potential for NLP methods that capture social meaning in language to help inform our understanding of society and its behaviors  \cite{kozlowskiGeometryCultureAnalyzing2019,bailey2022based,gargWordEmbeddingsQuantify2018}. As individuals who are supportive of existent critical literature, and who are still building methods that are inherently engaging in stereotyping, we must accept that our methods can be used not only to measure perceptions and self-presentations but also to apply them for nefarious purposes, such as racially-biased targeted advertising \cite{ali2019discrimination}. The ultimate ethical question that we must address, then, is whether we believe that our measurement strategies will ultimately do more harm than good.  

Our decision to submit the present work for consideration reflects our belief, after deliberation, that the potential benefits of our work outweigh the potential dangers. From a benefits perspective, we show that our work can help us understand the (mis)information environment on Twitter, and hope in the future to use these methods to better understand 1) how hegemonic voices are often given outsized attention on social media platforms, and 2) how attacks on marginalized communities changed along with the political climate in the United States.  From a misuse perspective, we do not believe that the methods we have developed here are more effective at targeting individuals based on stereotyped demographics than the tools already in existence elsewhere, tools which use orders of magnitude more compute data and power.  As such, while there are real dangers with being misclassified in a particular way online, we 1) emphasize throughout the paper that we focus only on how someone is likely to be perceived based on what is in their bio, and do not claim to be able to classify some ``true self'' of users, and 2) expect that if one wanted to engage in such a prediction activity, it could be done better with methods that focus specifically on this task.

A similar value judgement was applied to the other ethical question in our work: whether or not to publicly release parts of our data. Ultimately, restrictions on the Decahose \hl{have made it challenging to share data. However, given recent literature in the ICWSM community \cite{assenmacher2020end}, we believe the appropriate course of action is to release training data on a per-case basis, as potential dataset users contact the last author of this paper.} However, we believe that the release of our (fully deidentified) survey data and basic case study data does not serve any inherent risks for users in our dataset, and thus have opted to do so in the Github repository for this work. We believe that the utility of these data to the scientific community, both for replicability and extension, outweigh the dangers of data release. Of course, in all cases, data and methods usage is a continually evolving process, and we will re-evaluate this value judgement as often as is needed.
\begin{enumerate}

\item For most authors...
\begin{enumerate}
    \item  Would answering this research question advance science without violating social contracts, such as violating privacy norms, perpetuating unfair profiling, exacerbating the socio-economic divide, or implying disrespect to societies or cultures?
    \answerNo{As noted above, there are privacy concerns in our work that we have carefully considered and weighed relative to  the benefits of our work.}
  \item Do your main claims in the abstract and introduction accurately reflect the paper's contributions and scope?
    \answerYes{Yes}
   \item Do you clarify how the proposed methodological approach is appropriate for the claims made? 
    \answerYes{Yes}
   \item Do you clarify what are possible artifacts in the data used, given population-specific distributions?
    \answerYes{Yes}
  \item Did you describe the limitations of your work?
   \answerYes{Yes}
  \item Did you discuss any potential negative societal impacts of your work?
   \answerYes{Yes, see the Ethics Statement}
      \item Did you discuss any potential misuse of your work?
       \answerYes{Yes, see the Ethics Statement}
    \item Did you describe steps taken to prevent or mitigate potential negative outcomes of the research, such as data and model documentation, data anonymization, responsible release, access control, and the reproducibility of findings?
  \answerYes{Yes, see the Ethics Statement}
  \item Have you read the ethics review guidelines and ensured that your paper conforms to them?
   \answerYes{Yes}
\end{enumerate}

\item Additionally, if your study involves hypotheses testing...
\begin{enumerate}
  \item Did you clearly state the assumptions underlying all theoretical results?
    \answerNA{NA}
  \item Have you provided justifications for all theoretical results?
    \answerNA{NA}
  \item Did you discuss competing hypotheses or theories that might challenge or complement your theoretical results?
    \answerNA{NA}
  \item Have you considered alternative mechanisms or explanations that might account for the same outcomes observed in your study?
    \answerNA{NA}
  \item Did you address potential biases or limitations in your theoretical framework?
   \answerNA{NA}
  \item Have you related your theoretical results to the existing literature in social science?
    \answerNA{NA}
  \item Did you discuss the implications of your theoretical results for policy, practice, or further research in the social science domain?
    \answerNA{NA}
\end{enumerate}

\item Additionally, if you are including theoretical proofs...
\begin{enumerate}
  \item Did you state the full set of assumptions of all theoretical results?
    \answerNA{NA}
	\item Did you include complete proofs of all theoretical results?
   \answerNA{NA}
\end{enumerate}

\item Additionally, if you ran machine learning experiments...
\begin{enumerate}
  \item Did you include the code, data, and instructions needed to reproduce the main experimental results (either in the supplemental material or as a URL)?
    \answerYes{See the Github link in the first footnote}
  \item Did you specify all the training details (e.g., data splits, hyperparameters, how they were chosen)?
    \answerYes{See the main text as well as the appendix below}

     \item Did you report error bars (e.g., with respect to the random seed after running experiments multiple times)?
    \answerYes{Yes}
	\item Did you include the total amount of compute and the type of resources used (e.g., type of GPUs, internal cluster, or cloud provider)?
    \answerYes{Yes}
     \item Do you justify how the proposed evaluation is sufficient and appropriate to the claims made? 
    \answerYes{Yes. Specifically, we use a number of clear, competitive, and comparable baselines}
     \item Do you discuss what is ``the cost`` of misclassification and fault (in)tolerance?
    \answerYes{Yes, see the Ethics statement. }
  
\end{enumerate}

\item Additionally, if you are using existing assets (e.g., code, data, models) or curating/releasing new assets, \textbf{without compromising anonymity}...
\begin{enumerate}
  \item If your work uses existing assets, did you cite the creators?
    \answerYes{Yes.}
  \item Did you mention the license of the assets?
    \answerYes{To the best of our ability, we believe that nothing we have provided requires this, but we will update as necessary.}
  \item Did you include any new assets in the supplemental material or as a URL?
    \answerYes{Yes.}
  \item Did you discuss whether and how consent was obtained from people whose data you're using/curating?
    \answerYes{We provide details on the dataset, acknowledging that consent was approved in surveys as IRB requires. Social media users are non-consenting, which we acknowledge in our Ethics statement.}
  \item Did you discuss whether the data you are using/curating contains personally identifiable information or offensive content?
    \answerYes{Yes, see the ethics statement}
\item If you are curating or releasing new datasets, did you discuss how you intend to make your datasets FAIR?
\answerNo{In progress for the camera ready version}
\item If you are curating or releasing new datasets, did you create a Datasheet for the Dataset? 
\answerNo{In progress for the camera ready version.}
\end{enumerate}

\item Additionally, if you used crowdsourcing or conducted research with human subjects, \textbf{without compromising anonymity}...
\begin{enumerate}
  \item Did you include the full text of instructions given to participants and screenshots?
    \answerNo{We use a protocol from a prior work, which we do reference.}
  \item Did you describe any potential participant risks, with mentions of Institutional Review Board (IRB) approvals?
    \answerYes{Yes, our studies are IRB approved.}
  \item Did you include the estimated hourly wage paid to participants and the total amount spent on participant compensation?
    \answerYes{Yes.}
   \item Did you discuss how data is stored, shared, and deidentified?
   \answerYes{Yes.}
\end{enumerate}

\end{enumerate}

\appendix

\section{Appendix}
\subsection{Bios from the Twitter Decahose}
\label{sec:appendix_data_twitter}

\begin{table*}[t]
    \centering
    
    \begin{tabular}{|l|p{0.2\textwidth}|p{0.3\textwidth}||}
    \hline
        Dataset Portion & Number of distinct records in Twitter \\
        \hline
        \hline
        All raw records &  15,459,872 \\
        Initial training cut & 12,367,897\\
        Training dataset after cleaning & 3,534,903 \\
        Initial test cut & 3,091,975 \\
        Test dataset after cleaning & 1,546,001\\
        \emph{main} test dataset & 3,044,093 \\
        \emph{generalizability} test dataset & 395,583 \\
        Vocabulary (distinct phrases) & 22,516\\
        \hline
    \end{tabular}
    \captionsetup{justification=centering}
    \caption{Summary statistics for the identity-centric datasets we develop}
    \label{tab:datasum}
\end{table*}

\begin{table}[t]
    \centering
    \begin{tabular}{|l|p{0.11\textwidth}||}
    \hline
        identity & Number of times appeared \\
        \hline
        \hline
        she &  352,655\\
        her & 308,829 \\
        he & 144,845 \\
        him & 144,845 \\
        they & 353,4903 \\
        writer & 67,824 \\
        blm & 63,388 \\
        
        mixer streamer freak &  100\\
        published photographer & 100 \\
        sophomore & 100 \\
        micah 6:8 & 100 \\
        public health specialist & 100 \\
        britishindependence & 100 \\
        vikings fan & 100 \\
        
        \hline
    \end{tabular}
    \caption{Examples of the most and (some of the) least frequent identities in the Twitter dataset}
    \label{tab:dataex1}
\end{table}

The center column of Table \ref{tab:datasum} provides summary statistics for the Twitter bio data we construct. We begin with a sample of 15,459,872 distinct Twitter bios from users who posted a tweet in 2020 that was found in the Decahose, and who are specified as English-language users by the Twitter API.  In order to maintain a focus on culturally-salient identities, we limit the size of the vocabulary to identities used in at least 100 unique Twitter bios in the training set.  Further, because we are interested in associations between identities, we further remove Twitter bios that contain less than 2 identities. After these cleaning steps, our training and test data consists of 3,534,903 and 1,546,001 distinct bios respectively with 22,516 unique identities in the vocabulary. We then follow the approach outlined in the main text to produce the main test dataset and the generalizability test set.  Note that the size of each of these splits can be larger than the size of cleaned test dataset, because we can generate multiple instances from a given bio by randomly selecting different targets; i.e. we can generate multiple test instances out of each of profile description by selecting multiple pairs of $X^i_r$ and $X^i_t$.  

Finally, to provide further insight into the data, Table \ref{tab:dataex1} showcases the top 7 identities in terms of overall frequency in the training data and 7 of the least frequent identities to show that the tail still contains meaningful phrases.

\subsection{Single Identity Survey Data}\label{sec:appendix_data_survey}

\textbf{A complete, aggregated copy of this data is provided in the Github for this paper.} 
Our survey study was ruled exempt by the IRB at [REMOVED]. Each respondent rated between four and seven identities, and each identity was given to at least 3 respondents. Respondents were paid an average of \$12/hour. Respondents are from a convenience sample, as recent work has suggested that the cost efficiency of convenience samples does not necessarily impact data quality \cite{coppockGeneralizabilityHeterogeneousTreatment2018}.  The median age of our sample is 32. Of the 140 respondents, 88 reported their sex as female, 49 as male, and 4 noted other/did not provide.  Our sample, like Twitter, was overwhelmingly White; 105 (75\%) of the sample reported as White. 

We here provide two minor additional details on our survey data. First, it is of note that in contrast to prior work, we focus explicitly on priming respondents to think of social media users, asking, for example, ``If you saw [identity] in a social media biography, would you expect that individual to be'' and then provided a Likert scale ranging from (e.g.) ``Always [a] Democrat'' to ``Always [a] Republican.'' Second, we emphasize that other procedures, including tutorial materials, task details, and attention checks, follow the publicly available materials from \citet{joseph2020word}.

\subsection{Entire Bio Survey Data}

\textbf{A complete, aggregated copy of this data is provided in the Github for this paper.}

We use responses from approximately 730 Prolific respondents. Because of a temporary issue with the survey, some respondents were approved for pay without taking the study, we report demographics here on all respondents here because of this issue.  The median age of respondents was 37. Reported sex was more balanced than the sample for the single identity study, 51\% and 44\% of respondents reported a sex of female or male, respectively. As above, however, a majority of the sample (65\%) reported as White. Respondents were paid an average of \$12/hour. Full details are provided in the Appendix for the interested reader.

Two final notes are in order. First, with respect to the difference between the two surveys is that because we focus on individual user bios, we in this study ask respondents to assess the likelihood that the individual user who has this bio is, e.g., a Democrat or Republican, rather than asking (as above) about the probability that an individual who holds a given identity is, e.g., a Democrat or Republican. Second, we note that we drop the 5\% of respondents whose responses were furthest from all other respondents, on average, across all identities and dimensions (after standardizing measures across dimensions). This results in a final sample of 1,273 bios analyzed here and provided in the Github repository.

\subsection{Case Study}

\subsubsection{Replication with Data from the Decahose}\label{sec:appendix_newsguard}

We replicate findings using the case study dataset analyzed in the main text with a different dataset, broader in scope, drawn from the Decahose. 
\textbf{A complete, anonymized copy of this replication data is provided in the Github for this paper.} For each user, it contains the number of URLs they shared that link to low- and high-quality domains, as identified by NewsGuard, and projections of their bios onto the four dimensions of interest to the case study.

We begin with a sample of roughly 290,000 users who tweeted at least one NewsGuard URL in the decahose in 2020. Of this sample, 143,883 users were 1) still active when we recollected tweets in 2022, and 2) had a non-empty English-language bio. It is possible that our use of \texttt{langdetect} could bias our sampling of what is considered English language \cite{blodgettDemographicDialectalVariation2016}; as such, we manually evaluate it. To do so, we had three research assistants label 900 bios as English, not English, or Vague. The annotators had a Krippendorf's alpha agreement score of 0.81 and agreed on 90\% of the bios with the library. More specific to our case study, 93\% of the bios that \emph{langdetect} measured as English were also annotated as English by all the annotator. The main reason for mis-classification of the bio's language was that some bios consisted of phrases from multiple languages. We therefore believe that this step did not bias our results in any obvious way.

Of these remaining 143,883 users who matched our initial sampling criteria, a remaining 108,554, or 75.4\%, had at least 5 shares of URLs in NewsGuard, which we considered a minimum for estimating proportions.  The median user had 49 shares of high quality news website URLs and 3 shares of low quality news website URLs. In total, these users shared 11,735,521 links to NewsGuard domains we assess in our study. 

\begin{figure}
    \centering
\includegraphics[width=\linewidth]{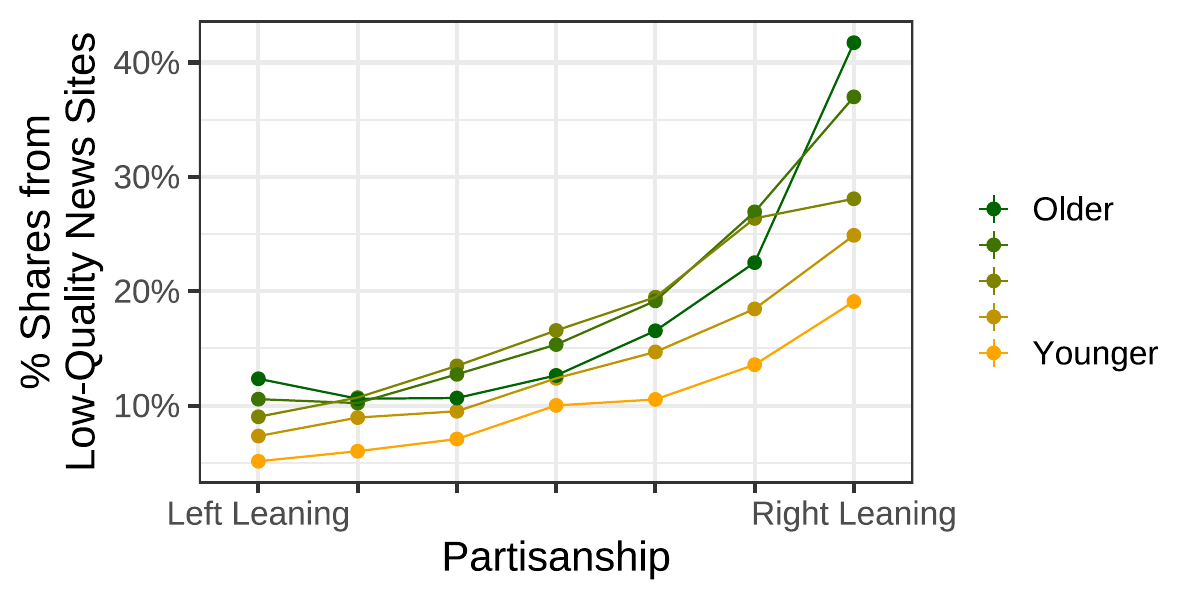}
    \caption{Replication of Figure~\ref{fig:age_partisanship}A) with a dataset drawn from the Decahose}
    \label{fig:age_repl}
\end{figure}

\begin{figure}
    \centering
\includegraphics[width=\linewidth]{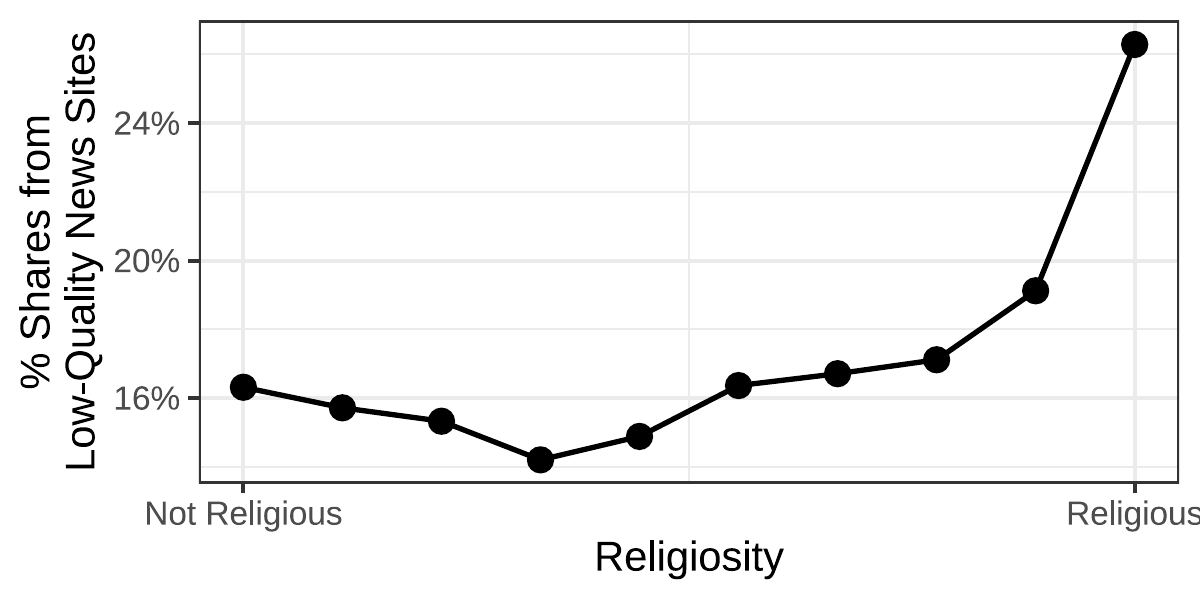}
    \caption{Replication of Figure~\ref{fig:age_partisanship}B) with a dataset drawn from the Decahose}
    \label{fig:rel_repl}
\end{figure}

Figure~\ref{fig:age_repl} replicates findings from Figure~\ref{fig:age_partisanship}A) in the main text, namely, observe a similar increase in the association between self-presented (older) age and misinformation sharing as we see an increase in right-leaning self-presentation. Figure~\ref{fig:rel_repl} replicates Figure~\ref{fig:age_partisanship}B).

\subsubsection{Descriptive Statistics Partisanship and Religiosity}

\begin{figure}
    \centering
\includegraphics[width=\linewidth]{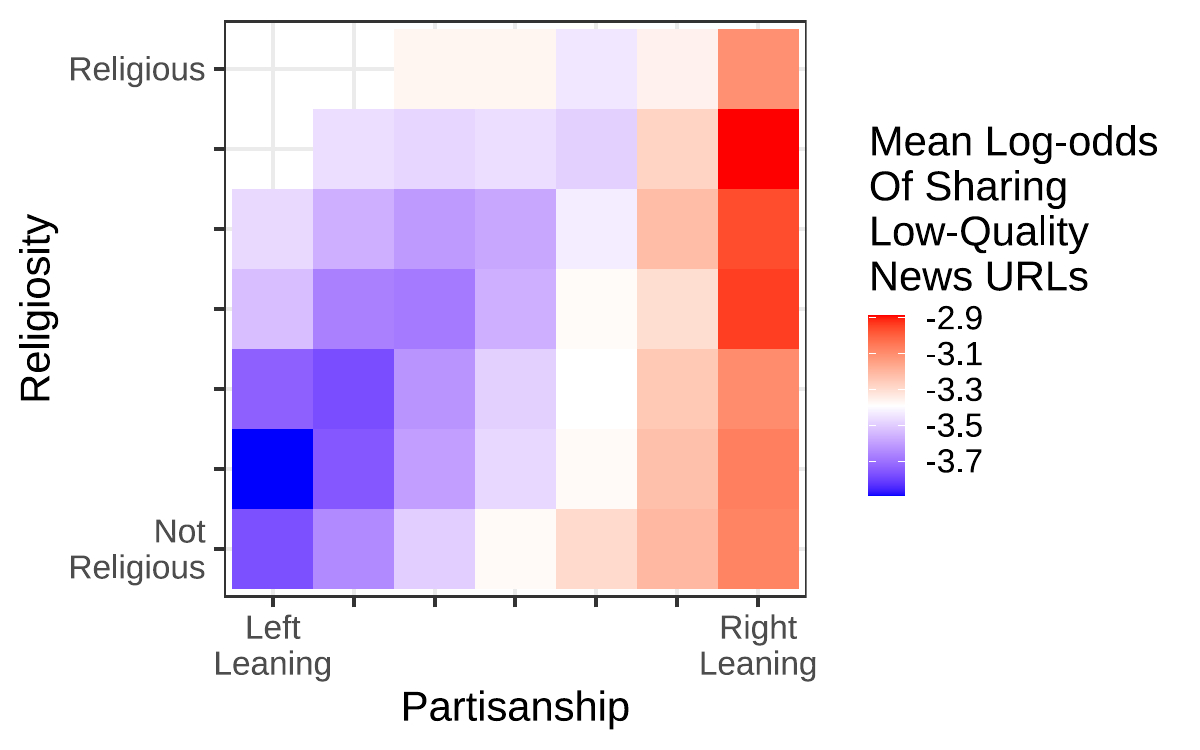}
    \caption{The mean log-odds of sharing a link from a low quality news site, relative to a high quality news site (color of each cell) for panel members with given levels of partisanship (x-axis) and religiosity (y-axis).}
    \label{fig:case_part_relig}
\end{figure}

Figure~\ref{fig:case_part_relig} shows a descriptive statistic that provides some evidence that religiosity and partisanship have an interactive effect on misinformation sharing. However, as noted in the main text, correlations between these variables make it challenging to assess the statistical significance of these variables relative to each other; in particular, there are few left-leaning accounts that also present as religious.

\subsection{Modeling}

\subsection{Distributional Semantic Models (Baselines)}\label{sec:appendix_model_dsm}

For all DSM baseline models except Sentence Bert, including BERT-base, RoBERTa-base and BerTweet-base we experimented using open-source implementations on Hugging Face transformers library \footnote{https://huggingface.co/docs/transformers/}. For the Sentence-Bert baseline, we used the \emph{mpnet-base} pre-trained model and the implementation given by open-source Sentence Transformers library.\footnote{https://github.com/UKPLab/sentence-transformers}

\subsubsection{Bio-only Model}\label{sec:model_w2v}

To select hyperparameters, we use 10\% of the training data as a validation dataset. The primary hyperparameter we tuned was whether to use a Skip-Gram or C-BOW model. We ultimately chose a Skip-Gram model for Twitter and a C-BOW model for Wikipedia, with the other hyperparameters as specified in the main text. Model training took under an hour on a personal laptop. We used the open-source implementation of \texttt{word2vec} in \texttt{gensim}\footnote{\url{https://radimrehurek.com/gensim/}} for our experiments.

\subsubsection{Fine-tuned SBERT}\label{sec:appendix_model_Fine-tuned}

        \begin{figure}[t]
            \centering
            \includegraphics[width=1\linewidth]{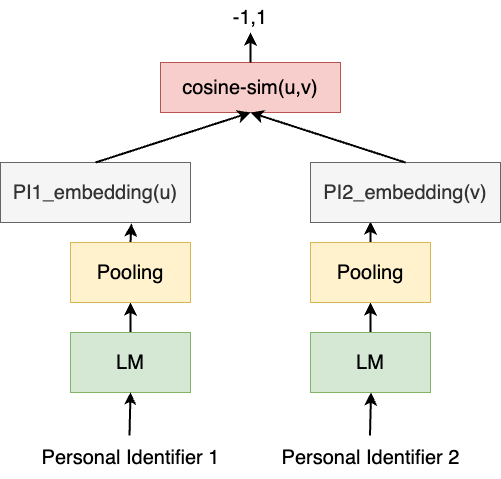}
            \caption{Training procedure for contrastive learning with regression objective function.}
            \label{fig:loss}
        \end{figure}

We here briefly provide additional intuition for our Fine-tuned SBERT model. Given a dataset of pairs of identity phrases with a label denoting the anchor-positive pair or anchor-negative pairs, we input the pair into the pipeline shown in Figure \ref{fig:loss} and extract the latent embeddings of each of the identities. Then we calculate the cosine similarity of the embeddings and backpropagate the mean squared error loss with the label through the network. In this work we fine-tuned a \emph{mpnet-base} Sentence Bert model since it had the same model size as a BERT-base and the pre-trained model was available online.

\subsubsection{Further details on Baseline Models}

For all DSM baseline models except Sentence Bert, including BERT-base, RoBERTa-base and BerTweet-base we experimented using open-source implementations on Hugging Face transformers library \footnote{https://huggingface.co/docs/transformers/}. For the Sentence-Bert baseline, we used the \emph{mpnet-base} pre-trained model and the implementation given by open-source Sentence Transformers library.\footnote{https://github.com/UKPLab/sentence-transformers}.

\subsubsection{Other Baseline Models Considered}
\label{sec:appendix_models_others}

% \begin{figure}[t]
%     \centering
%     \includegraphics[width=1\linewidth]{figures/performance-others.png}
%     \caption{Average rank, log softmax score and top 1 percent accuracy of the target PI given by  Bertweet model and all of the projection models derived from main proposed models}
%     \label{fig:performance_others}
% \end{figure}

  In addition to the three baseline models discussed in the text, we also experimented with a pair of other sensible options. 
  
  First, we expected that a DSM pretrained on Twitter would be a strong baseline to compare to, and thus  experimented with additional models pre-trained specifically on Twitter data \cite{nguyen-etal-2020-bertweet}. We use the fine-tuned BERT model on Twitter data proposed by \cite{nguyen-etal-2020-bertweet}. They propose a BERT-base model fine-tuned using a corpus of 850M English Tweets. However, we find that model performance was no better than the other, more widely used baseline DSMs we proposed in the main experiments.
  
  Second, it seemed reasonable that by first restricting a baseline DSM to known dimensions of social meaning, we could improve their performance. Consequently, we considered baselines where we first projected down all baseline models into the core dimensions of meaning noted by \citet{joseph2020word} before the evaluation tasks. In both cases, however, our intuitions did not match empirical reality. These models failed to outperform the baselines used in the main text, and thus we restrict our analysis to the baselines discussed in the main text.

% \subsubsection{Projection Based Models}
% For all the main proposed models, we first project down all the embeddings into the core dimensions of meaning noted by \citet{joseph2020word} and then measure the similarity scores and rankings according the evaluation tasks. We also didn't focus on these models since it's performance was not noticeable. The predictive performance results for these models are shown in Figure \ref{fig:performance_others}.

\subsubsection{Generating Embeddings for the Predictive Experiment}

In order to build inputs to the network, since $X^i_r$ is a list of personal identifiers, to calculate the latent embedding $L^i_r$ for it, depending on the model, we follow different procedures. For the \emph{Bio-only} model, we simply measure the average latent vector of all phrases in $X^i_r$ according to \eqref{eq:w2vavg}. For the Fine-tuned models, as well as the baseline contextualized language models discussed below, we stitch the words in $X^i_r$ together with comma and create a sentence $S^i_r$. We then measure $L^i_r$ according to Equation \eqref{eq:lmavg}. Equivalently, this means that for the BERT based models we take the embedding of [CLS] token for pooling and for the Sentence Bert based models we follow the original work and take the average of all token embeddings.

\begin{gather}
L^i_r = \sum_{v \in X^i_r}(v)/|X^i_r| \label{eq:w2vavg} \\
L^i_r = Pooling(LM(S^i_r)) \label{eq:lmavg} 
\end{gather}  

\subsection{Projection to Social Dimensions}
\label{sec:appendix_projection}

In order to project a piece of text (either a full bio or a single identity) onto a specific dimension, we have to first define the end-points of that dimension. Table \ref{tab:olddims} and \ref{tab:newdims} outline the original and in-domain dimension end-points that we talk about in this paper, in particular in our Entire bio projection evaluation.
To generate embeddings for an end-point, we assume each end-point is an instance in $X$ and follow the approach outlined above to generate embeddings for each end-point of the dimension $d$ and call them $E_d^1$ and $E_d^2$ respectively. Having the embedding of both poles of the dimension, we calculate the difference vector according to Equation \ref{eq:diffvec}, and calculate the embedding of the target text using the same approach to a vector $L^i$. Then we follow the projection approach outlined in \citep{ethayarajhUnderstandingUndesirableWord2019} to normalize all vectors and then calculate the projection value of $L^i$ onto dimension $d$ according to equation \ref{eq:prj}.

\begin{equation}
\label{eq:diffvec}
    V^d = E_2^d - E_1^d
\end{equation}

\begin{equation}
\label{eq:prj}
    P^i_d = \frac{V^d.L^i}{||L^i||}
\end{equation}

\begin{table*}[t]
    \centering
    
    \begin{tabular}{|l|p{0.2\textwidth}|p{0.3\textwidth}||}
    \hline
        Dimension & End point 1 & End point 2 \\
        \hline
        Age & young, new, youthful &  old, elderly, aged\\
        \hline
        Partisanship & democratic party supporter, left-leaning, democrat & republican party supporter, right-leaning, republican\\
        \hline
        Religion & atheistic, agnostic, non-believing, skeptical & religious, faithful, christian, believe in lord\\
        \hline
        Politics & music, sports, culture, tech & politics, political, democrat, republican\\
        \hline
        Gender & mother of, grand mother & father of, grand father \\
        \hline
    \end{tabular}
    \captionsetup{justification=centering}
    \caption{Dimension endpoints for ``Original'' dimension endpoint selection method}
    \label{tab:olddims}
\end{table*}

\begin{table*}[t]
    \centering
    \footnotesize
    \begin{tabular}{|l|p{0.35\textwidth}|p{0.35\textwidth}||}
    \hline
        Dimension & End point 1 & End point 2 \\
        \hline
        Age & 15 years old, 18 years old, sophomore in college, student at, umich22, 18, 21 &  retired person, I’m old, 50 years old, 65 years old, 61yr old, grandparent of, old man, old woman, grandma to, grandpa to, tenured, long career\\
        \hline
        Partisanship & pro socialism, liberal democrat, never trump, proud democrat, vote blue no matter who, \#resist, \#voteblue, \#nevertrump, left leaning, \#democraticdownballot, \#notvotebluenomatterwho, \#bidenharris, \#resist, \#bluewave, \#democraticsocialist & right leaning, trump won, never biden, fuck biden, \#maga, \#kag, Trump conservative, conservative and America First, proud Trump supporter, trump fan, \#MAGA Republican, constitutional conservative patriot, \#trump2024\\
        \hline
        Religion & atheist, nonbeliever, proud atheist, totally secular, \#cancelreligion & Catholic, jesus christ, follower of christ, priest, lover of jesus, christian episcopalian, jesus loving christian, john 3:16, gospel of the lord jesus christ, minister at united church, christ-follower, god first, isaiah 55:6, woman of faith, man of faith\\
        \hline
        Gender & sister, wife, mother, Proud Mama and Wife, grandmother of, mother of one, mama of one, wife of, Loving Wife, she, her, hers & husband to, brother, husband, father, grandfather of one, father of one, Loving husband, he, him, his, son, brother, brother-in-law, uncle, nephew \\
        \hline
    \end{tabular}
    \captionsetup{justification=centering}
    \caption{Dimension endpoints for ``In-Domain'' dimension endpoint selection method}
    \label{tab:newdims}
\end{table*}

\subsection{Error Analysis of The Predictive Task}
\label{sec:appendix_err_analysis}

Our understanding of the proposed models is improved by studying where errors occur. Here, we briefly present both quantitative and qualitative reflections on the major sources of error for the Bio-Only and Fine-tuned SBERT models. At a high level, we find that pre-training helps the Fine-tuned SBERT model on predictions requiring knowledge of phrase composition (e.g. that ``mother'' and ``mother of two'' convey similar  meanings), but appear to cause it to struggle on infrequent identities, which the Bio-only model is better able to capture meanings of from the bio data alone.

\begin{figure}[t]
        \centering
        \includegraphics[width=1\linewidth]{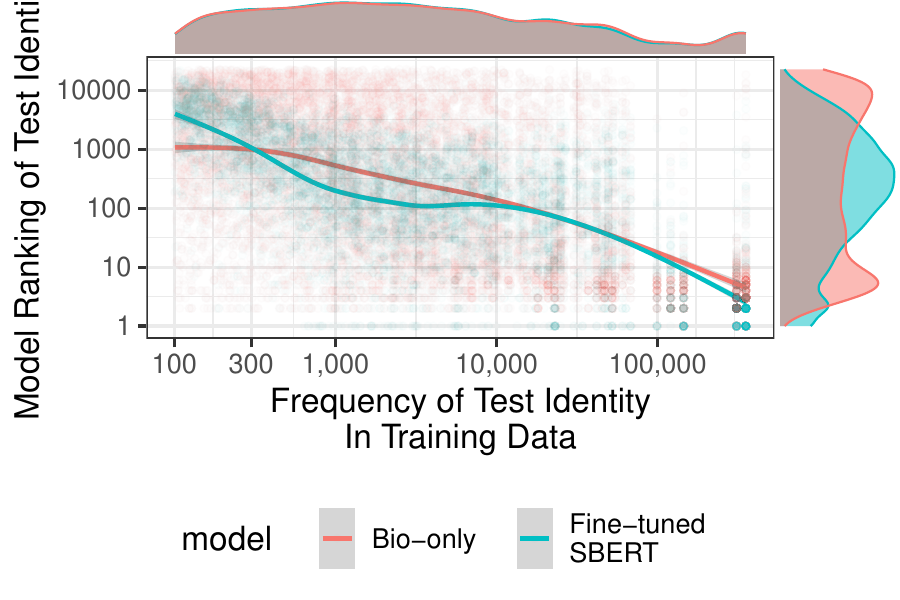}
        \caption{
            Average rank (y-axis) versus the frequency of that phrase in training dataset (x-axis) for 10K sample test points (each overplotted point) for the Bio-only and Fine-tuned SBERT models. Fit line is a generalized additive model; marginal densities are also displayed.
        }
        \label{fig:rankfreq}
    \end{figure}

Quantitatively, Figure~\ref{fig:rankfreq} shows that both models performed best, and roughly equally well, on the most frequent identities, but that differences appeared in how the models fared elsewhere.
The Bio-only model's ranking distribution (the marginal density plot on the right-hand side of Figure~\ref{fig:rankfreq}) was bimodal, with a large number of high (poor performance) and low (strong performance) ranks for test points. Perhaps unsurprisingly, we find qualitatively that the poor performance of the Bio-only model relative to the Fine-tuned SBERT model largely came from an inability  1) to learn from compositional identities or 2) to leverage relevant external knowledge. These issues seemed to impact the model most for moderately frequent target identities, those appearing between 300-10,000 times in the training data. With respect to 1), for example, when provided the Twitter bio ``mother of two, restaurant owner, partly retired, hockey coach''\footnote{This bio has been modified to protect user privacy}, the Bio-only model ranks the correct held-out identity, ``wife,'' among the least likely. In contrast, the Fine-tuned SBERT model correctly ranks ``wife'' in the Top 1\%. The core difference is that the Fine-tuned SBERT model, but not the Bio-only model, leverages the gender stereotype implied by the ``mother'' portion of the phrase ``mother of two.''  With respect to 2), there were several cases where external knowledge from the pre-trained model benefited the Fine-tuned models.  For example, the Fine-tuned models, but not the Bio-only models, were able to recognize the similarity between the identities ``follower of ISKSON'' (a Hindu religious organization)  and ``proud Hindu.'' Both of these were relatively infrequently used.

In contrast, relative to the Bio-only model, Fine-tuned models struggled with the \emph{most infrequent} identities, in particular the roughly 18\% of identifiers in the test set that occurred fewer than 300 times in the training data. In these cases, as in prior work entity-centric domain adaptive work \cite{fieldEntityCentricContextualAffective2019}, the Fine-tuned models seemed to rely too heavily on knowledge from the pre-trained model and not enough to domain-relevant context. In contrast, the identity-centric model seemed to benefit on the prediction task from overfitting to stereotypical knowledge for these rarer phrases. %This is surprising, but is similar in certain ways to the findings of \citet{wolfe-caliskan-2021-low}. 
The Fine-tuned models also struggled when presented with identities, such as Twitter-specific acronyms, that were likely rare in the DSM data, but more frequent on Twitter. Here, pre-training seemed to induce noise, leading the Fine-tuned models to predict somewhat randomly.

%\subsection{Additional Figures}

% \begin{figure}
%     \centering
%     \includegraphics[width=\linewidth]{figures/age_part_2.pdf}
%     \caption{Replication of results in Figure~\ref{fig:age_partisanship} except using n=10 bins for both partisanship and age}
%     \label{fig:appendix_bin_2}
% \end{figure}

% \begin{figure}
%     \centering
%     \includegraphics[width=\linewidth]{figures/age_part_3.pdf}
%     \caption{Replication of results in Figure~\ref{fig:age_partisanship} except using n=5 bins for partisanship and n=10 bins for age}
%     \label{fig:appendix_bin_3}
% \end{figure}

% \begin{figure}
%     \centering
%     \includegraphics[width=\linewidth]{figures/gender_pres.pdf}
%     \caption{Expected proportion of low-quality news shares with n=10 bins for projections onto gender. Error bars are 95\% CIs using normal approximations.}
%     \label{fig:appendix_gender}
% \end{figure}

% \begin{figure}
%     \centering
%     \includegraphics[width=\linewidth]{figures/rel_pres.pdf}
%     \caption{Expected proportion of low-quality news shares with n=10 bins for projections onto religiosity. Error bars are 95\% CIs using normal approximations.}
%     \label{fig:appendix_religiosity}
% \end{figure}

\end{document}